\documentclass[submission,copyright,creativecommons]{eptcs}
\providecommand{\event}{FMAS 2024} 
\usepackage{underscore}
\usepackage{graphicx}
\usepackage{xcolor}
\usepackage{wrapfig}
\usepackage{subfigure}
\usepackage{enumitem}
\usepackage{multirow}
\usepackage{tcolorbox}
\usepackage{footmisc}
\usepackage{lineno}
\usepackage{tikz}
\usepackage{amsfonts}
\usepackage{amsmath}
\usepackage{mathtools}
\usepackage{setspace}
\usepackage[linesnumbered,ruled,vlined]{algorithm2e}
\usepackage{tabularx}
\title{Model Checking for Reinforcement Learning in Autonomous Driving: One Can Do More Than You Think!}
\author{Rong Gu
\institute{M{\"a}lardalen University\\ V{\"a}ster{\aa}s, Sweden}
\email{rong.gu@mdu.se}}
\def\titlerunning{Model Checking for Reinforcement Learning in Autonomous Driving}
\def\authorrunning{Rong Gu}
\begin{document}
\maketitle
\begin{abstract}
Most reinforcement learning (RL) platforms use high-level programming languages, such as OpenAI Gymnasium using Python. These frameworks provide various API and benchmarks for testing RL algorithms in different domains, such as autonomous driving (AD) and robotics. These platforms often emphasise the design of RL algorithms and the training performance but neglect the correctness of models and reward functions, which can be crucial for the successful application of RL. This paper proposes using formal methods to model AD systems and demonstrates how model checking (MC) can be used in RL for AD. Most studies combining MC and RL focus on safety, such as safety shields. However, this paper shows different facets where MC can strengthen RL. First, an MC-based model pre-analysis can reveal bugs with respect to sensor accuracy and learning step size. This step serves as a preparation of RL, which saves time if bugs exist and deepens users' understanding of the target system. Second, reward automata can benefit the design of reward functions and greatly improve learning performance especially when the learning objectives are multiple. All these findings are supported by experiments.
\end{abstract}
\section{Introduction}\label{sec:introduction}
With the advance of hardware and artificial intelligence (AI), \textit{Autonomous Driving (AD)} has become more and more realistic around us.
However, although automotive companies are running road tests for their AD vehicles over millions of miles a year, accidents still keep occurring~\cite{zhang2022finding}, like crashes involving Tesla’s driver-assistance system \cite{teslaAcc}, and a fatal crash caused by a self-driving car of Uber \cite{uberAcc}.
Such accidents damage people's confidence in AD dramatically as the public usually cannot accept even a single accident caused by an AD vehicle.
One of the reasons is that people even experts cannot fully understand why AD vehicles behave in a certain way as the AI components are different from conventional hardware and software systems and they can be unpredictable. 
Without knowing the reason behind an AD vehicle's every action, it is impossible to gain trust in the machine.
Formal methods (FMs) are widely accepted for their ability to analyse safety-critical systems with mathematics-based methods. In recent years, scientific studies and projects have been conducted where FMs play an important role in providing safety assurance on AD systems \cite{dpac}\cite{satisfies}\cite{gu2022correctness}\cite{sanchez2022foresee}\cite{yang2023reinforcement}\cite{garg2024learning}.
However, FMs have limits when adopted in AD systems, such as the usability barrier due to the complex mathematical models and scalability due to the notorious state-space explosion.
Another problem with using FMs in AD systems is the lack of tools that can provide visualisation of models, counterexamples, and analytical results.
Thanks to our previous work, two state-of-the-art tools in both FMs and AD research communities have been integrated, namely \textit{CommonUppRoad} \cite{gu2024commonupproad}.
This new tool combines the model checker UPPAAL \cite{larsen1997uppaal} with CommonRoad \cite{althoff2017commonroad}, an open-source toolset for AD development, testing, and visualisation.
Users of CommonUppRoad can load and configure an AD scenario and specify the planning goal by programming it in Python.
The tool then converts everything into a formal model that is recognisable and verifiable by UPPAAL.
Although UPPAAL is well-known for its ability of symbolic and statistical model checking of timed automata, the latest version of UPPAAL\footnote{UPPAAL 5 on uppaal.org} also provides functions for controller synthesis.
Specifically, UPPAAL can compute a so-called \textit{strategy} that controls which transitions (aka, actions) to choose at each of the states.
In this way, the model state space is restricted by the strategy and the exhaustive-search-based synthesis in UPPAAL guarantees that the strategy fulfils properties, e.g., safety.
Additionally, UPPAAL also provides reinforcement learning (RL) algorithms to optimise a strategy.
For example, a safe strategy can permit multiple actions at a state, but only one of them moves the AD vehicle towards the goal.
RL can capture this action by accumulating rewards of state-action pairs and eventually control the AD vehicle to always choose the actions with the highest reward, i.e., the goal-reaching actions.
Additionally, the fact that the learning is performed under the control of a safe strategy ensures the learning process as well as the result are still safe, and this is not given by pure RL.
One of the barriers to using UPPAAL in the AD domain is that the strategies of UPPAAL are not visualised properly. One cannot get to know the moving trajectories of an AD vehicle under the control of a UPPAAL strategy.
CommonUppRoad compensates for this disadvantage by leveraging CommonRoad's ability to visualise the driving scenarios and its rich database of real-world traffic scenarios.
Despite all the advantages, CommonUppRoad has unsolved problems. First, the synthesis of safe strategies is done via an exhaustive search of the state space. This is very similar to the symbolic model checking of UPPAAL. Therefore, the state-space explosion of symbolic model checking also exists in the synthesis of safe strategies.
According to our experiments \cite{gu2024commonupproad}, the computation time rises exponentially when the maximum execution time of the AD vehicle increases linearly.
Second, the motion model of AD vehicles in CommonUppRoad is not precise enough to represent continuous actions. For example, turning in CommonUppRoad only has two speeds, i.e., $\pm 0.1\ d/s$, which is not faithful to the vehicular dynamics in CommonRoad.
Third, when the tool returns ``no result'', it means the state space does not contain a sequence of actions that satisfies the safety property, such as never colliding.
However, it is hard to analyse where the bug originates because it can be caused by the sensor error or the control logic of the strategy.
Last, besides the safety guarantee, model checking (MC) does not benefit RL in CommonUppRoad. However, there are many aspects that the former can do for the latter. For example, the design of the reward function is a dominant factor influencing learning efficiency and effectiveness.
However, there is a lack of an automatic validation framework for reward functions in the context of AD \cite{abouelazm2024review}.
The author believes MC can contribute to establishing such a framework.
This paper introduces how MC benefits RL in the AD context.
First, the author proposes new model templates for AD vehicles and their driving scenarios in the new CommonUppRoad framework.
By using the new model templates, analysis of sensor accuracy and a quick check on the existence of solutions can be conducted prior to RL.
This step is highly beneficial but neglected by most studies.
Second, the new model templates allow multiple timesteps of sensing and decision-making, as well as finer granularities of action discretisation, which can be taken as \textit{continuous} actions in practice.
Last, the author shows how symbolic and statistical model checking benefits the design and evaluation of reward functions.
Statistical model checking and the corresponding model templates enable statistical analysis of probabilistic models, which are important for AD design and testing in uncertain environmental models.
In a nutshell, the contributions of this paper are as follows:
\begin{itemize}
    \item New model templates supporting ``continuous actions'' – a fine granularity of discretisation, and the corresponding model conversion from CommonRoad to UPPAAL.
    \item Model pre-analysis before RL. The analysis shows the data accuracy that RL can tolerate, suitable periods for sensing and decision-making, and the possible existence of an optimal motion plan. A safety shield can also be automatically generated prior to RL such that the state-space exploration of RL is restricted to safe regions.
    \item Reward automata. This model and the corresponding analysis benefit RL designers in understanding the system model and reward functions. With reward automata, the learning performance can be greatly improved especially when the learning objectives are multiple.
\end{itemize}
The remainder of the paper is organised as follows. Section \ref{sec:preliminary} introduces CommonUppRoad, which is the foundation of this study. Section \ref{sec:model} describes the new model templates before Section \ref{sec:mc4rl} introduces how model checking strengthens reinforcement learning. Section \ref{sec:experiment} presents the experiments and results. Section \ref{sec:related} presents the related work and Section \ref{sec:conclusion} concludes the paper and introduces future work.
\section{Preliminary}\label{sec:preliminary}
In this section, the author briefly introduces the aspects of CommonUppRoad that are necessary for understanding this paper. Interested readers are referred to the literature \cite{gu2024commonupproad} for a detailed description of the tool including experimental results.
\subsection{Models of AD Vehicles and Environment}
\begin{figure}[t]
    \vspace{-0.0cm}
    \setlength{\abovecaptionskip}{-0.0cm}   
    \setlength{\belowcaptionskip}{-0.2cm} 
    \centering
    \includegraphics[width=0.85\textwidth]{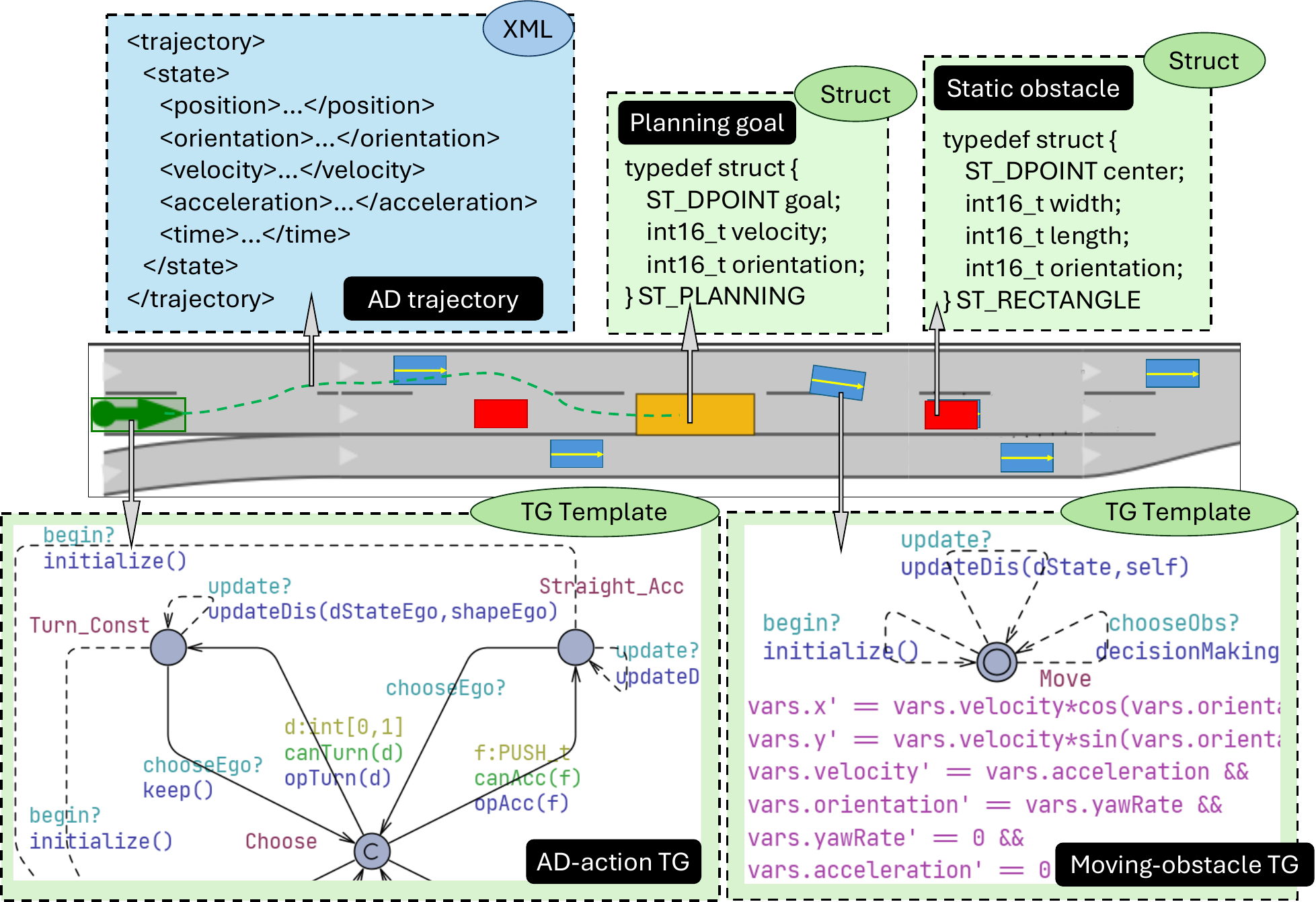}
    \caption{CommonUppRoad, a platform for model-checking-based AD motion planning, verification, and visualisation. Figure adopted from the literature \cite{gu2024commonupproad}.}
    \label{fig:commonupproad}
\end{figure}

In CommonUppRoad, users can load and configure an AD scenario that is stored as an XML file in CommonRoad \cite{althoff2017commonroad}, one of the fundamental tools of CommonUppRoad.
Essentially, a scenario contains a network of roads, static vehicles and traffic signs on the roads, a planning goal, and a group of moving vehicles that can follow predefined trajectories or behave reactively to other vehicles.
Static elements in the scenario, such as static obstacles, are presented as a group of XY coordinates in the XML file, representing their shapes (e.g., rectangle) and positions in a 2D environment.
The states of moving vehicles consist of their positions, orientations, velocities, and accelerations at each of the time points until the maximum time.
The scenario XML file stores moving trajectories as sequences of states that are sorted from the beginning to the maximum time.
An example of moving trajectories is shown in Fig.\ref{fig:commonupproad}.

CommonRoad provides Python functions for visualising and parsing the XML file.
CommonUppRoad calls these functions in the model conversion from AD scenarios to formal models, that is, timed games (TG) in UPPAAL.
As a class of formal model, TG has syntax and semantics.
Briefly, the syntax is the formal presentation of the model structure and the semantics is an interpretation of the syntactical model.
Fig.\ref{fig:commonupproad} has the snippets of two TG, which depict their syntactic presentations.
The box on the bottom left is the AD-action TG. Although it is only partially shown in the figure (for the sake of page limit), one can see that the turning action, an edge from location \texttt{Choose} to location \texttt{Turn\_Const}, is discrete.
In UPPAAL, integers can be assigned to edges via a \textit{select} statement, distinguishing syntactically the same edge at the semantic level. 
For example, in the AD-action TG, a select statement (\texttt{d:int[0,1]}) associates an integer \texttt{d}, whose value can be 0 or 1, to the edge from location \texttt{Choose} to location \texttt{Turn\_Const}, meaning that turning has two angular speeds.
Similarly, other actions that are syntactically represented by one edge can have multiple semantic transitions.
Controller synthesis means to choose from those transitions at each of the states such that the TG can satisfy properties regardless of how the environment's actions, that is, the dashed edges, are performed.
For example, the moving-obstacle TG in Fig.\ref{fig:commonupproad} only has one location and three dashed edges, meaning that after initialisation, moving obstacles keep changing their state variables continuously, updating their discrete variables and making decisions periodically.
If the planning goal is to reach a destination, then motion planning here means to choose actions at states of the AD-action TG such that the destination is reachable no matter how the actions in the moving-obstacle TG are taken.
\subsection{Motion Planning Methods and Visualisation}
CommonUppRoad has two kinds of motion-planning methods, i) exhaustive-search-based synthesis (ESS), and ii) reinforcement-learning-based synthesis (RLS).
\subsubsection{Exhaustive-Search-Based Synthesis (ESS)}
As synthesis is about finding a combination of actions of the AD vehicle, a natural method is to exhaustively explore the model state space and collect the desired execution traces, i.e., sequences of state-action pairs.
ESS follows this idea but the problem is that timed automata have infinite and uncountable states because of the continuous variable, \textit{time}.
UPPAAL uses \textit{zones} to represent the infinite values of time such that the symbolic state space of timed automata is finite and countable \cite{bengtsson2003timed}.
ESS uses the symbolic state space and the exploration is on the fly, that is, checking the property while exploring the state space simultaneously.
Specifically, CommonUppRoad uses the following query in UPPAAL, where \texttt{A[]} means all the states of all the traces in the state space, and functions \texttt{collide()} and \texttt{offroad()} are functions for judging if a collision or going off the road ever happens, respectively.
Therefore, Query (\ref{query:safe}) aims to collect all the state-action pairs where those two functions return \textit{false}.
\begin{equation}\label{query:safe}
    \texttt{strategy safe = control: A[] !collide() $\&\&$ !offroad()}
\end{equation}
When Query (\ref{query:safe}) is executed, UPPAAL explores the entire symbolic state space of the model of the AD vehicle and environment, checks if \texttt{collide()} or \texttt{offroad()} ever returns \textit{false}, and stores the execution traces where the check passes.
If a strategy is synthesised, it is represented as a function mapping actions to each of the states.
Since the state-space exploration is exhaustive, ESS is \textit{sound} and \textit{complete}, that is, the synthesised strategy must be correct, i.e., free of colliding and offroad situations (soundness), and if a correct strategy exists, the method must be able to find it (completeness).
Henceforth, we call strategies computed via ESS safe strategies.
A safe strategy is \textit{permissive} as it contains all the safe state-action pairs, including the inefficient ones.
For example, when an AD vehicle needs to turn left at an intersection, a permissive strategy would allow the AD vehicle to wait unnecessarily long, e.g., close to the maximum execution time, and then turn.
Therefore, permissive strategies can be optimised.
\subsubsection{Reinforcement-Learning-Based Synthesis (RLS)}
RLS explores the model state space randomly via simulations instead of exhaustive exploration.
Note that states in RLS are not symbolic anymore as they have concrete values of the state variables, including time.
Scores of the state-action pairs are computed through a reward function and accumulated during the random exploration.
After user-defined rounds of simulation (aka, learning episodes), learning finishes with a score table containing the pairs and their scores.
Following the control of a learned strategy means always choosing the action with the highest/lowest score at each of the states.
The advantage of RLS is that state-space explosion does not exist as the exploration is not exhaustive now, but the method is neither sound nor complete.
We need methods to provide correctness guarantees on the learning results.
CommonUppRoad uses safety strategies synthesised by UPPAAL running Query (\ref{query:safe}) to achieve this goal.
Specifically, RL executes the following query in UPPAAL, where \texttt{maxE(reward)} means the learning objective is to maximise rewards, \texttt{MAXT} is the maximum time of one learning episode, \texttt{<>} is a temporal operator meaning the existence of a state in any of the traces, \texttt{goal()} is a function to judge if the AD vehicle reaches the destination, and \texttt{under safe} means the random selection of state-action pairs must be within the pair set of strategy \texttt{safe}. 
Hence, the safe strategy serves as a safety shield for learning and must be synthesised prior to running Query (\ref{query:reachSafe}), and the latter aims to sample and compute the scores of the pairs from traces that have a state reaching the destination and the state space must be restricted by the safe strategy (aka, safety shield).
\begin{equation}\label{query:reachSafe}
    \texttt{strategy reachSafe = maxE(reward)[<=MAXT]: <> goal() under safe}
\end{equation}
Strategies can be visualised in CommonUppRoad as animated moving trajectories\footnote{Trajectory animations are posted online: sites.google.com/view/commonupproad/experiment}.
Specifically, CommonUppRoad simulates the model by using the following query, where \texttt{x}, \texttt{y}, and \texttt{velocity} are the representative variables that are in the state structure of trajectories in Fig.\ref{fig:commonupproad}.
Query (\ref{query:simulate}) randomly simulates the model under the control of strategy \texttt{reachSafe}, samples values of the variables in the curly brackets, and stores them in a log file, which is parsed in CommonUppRoad for visualisation. 
Strategy visualisation is another important feature that CommonUppRoad provides because there was no easy way to visualise UPPAAL strategies as moving trajectories.
\begin{equation}\label{query:simulate}
    \texttt{simulate [<=MAXT]\{x, y, velocity, ...\} under reachSafe}
\end{equation}
\subsubsection{Unsolved Problems in CommonUppRoad}
\textbf{Modelling of Continuous Actions}. As shown in Fig.\ref{fig:commonupproad}, the AD-action TG only has discrete actions. For instance, turning only has two possible angular speeds in CommonUppRoad but the vehicle dynamics in CommonRoad can have continuous actions, e.g., a continuously changing angular speed represented by a real number.
However, formally modelling continuous actions is not trivial. 
Once the model includes continuous variables, safe shields as Query (\ref{query:safe}) are not supported anymore unless those variables are hybrid clocks.
%
%
%
Hybrid clocks are special variables in UPPAAL. 
They are continuous and their changing rates are described by ordinary differential equations (ODE), but they cannot be used in guards (i.e., boolean expressions on edges) or invariants (i.e., boolean expressions on locations) as the values of hybrid clocks are not supposed to change the model behaviour. 
In other words, hybrid clocks are abstracted away from symbolic analysis, and thus UPPAAL can still use zones to form a symbolic state space of the model.
Therefore, although moving obstacles have nonlinear and continuous dynamics, their variables are modelled as hybrid clocks in UPPAAL and thus CommonUppRoad can still synthesise safety shields for reinforcement learning.
Modelling AD-action TG is not the same. 
%
%
%
If one models the AD vehicle's actions as hybrid clocks and ODE, they would not be taken into the symbolic state space and thus safety-shield synthesis would have no actions to learn from.
This is not the problem of the tool, UPPAAL, but rather a theoretical limit.
Models with both discrete and continuous components are hybrid automata, and the reachability problem of hybrid automata is undecidable in general \cite{henzinger1995s}.
Hence, there is no model checker that supports exhaustive verification of hybrid automata so far.
Hybrid clocks provide a way to model continuous actions but one needs methods to represent those continuous actions symbolically.
\noindent\textbf{Model pre-analysis before RL}. Like most studies of AD motion planning, CommonUppRoad attempted to synthesise an AD controller without analysing the model itself.
However, there are at least two questions before synthesis starts: i) are the data perceived by the digital controller of an AD vehicle accurate enough, and ii) does a valid motion plan exist in the model state space?
Question i) comes from the fact that the controller of an AD vehicle is a piece of software that periodically reads data about the surrounding environment and sends signals to control the vehicle.
As a piece of software, no matter how accurate the sensors are, it inevitably truncates real numbers to floating-point numbers, which have a limited length of digits.
Additionally, sensors cannot be perfect. When a new measuring method is introduced into an AD system, one may want to investigate the intervals of sensor errors before running motion planning. 
%
%
Question ii) comes as a following concern after question i). Given the current sensor error and motion primitives (aka, the atomic motions used in RL), does a valid solution exist?
If one can get a negative answer to this question quickly, there is no need to run RL at all. 
%
Therefore, a model analysis before RL can be greatly beneficial.
%
%
%
\section{New Model Templates of AD Vehicles and Environment}\label{sec:model}
Before introducing what model checking can do for reinforcement learning in the context of AD vehicles, the author describes the new model templates of CommonUppRoad in this section.
The new model templates allow symbolic and statistical model checking, which play the fundamental role in the model pre-analysis and reward automata design (Section \ref{sec:mc4rl}).
\subsection{Model Templates of AD Vehicles}\label{sec:model-templates}
\begin{figure}[t]
\vspace{-0.0cm}
\setlength{\abovecaptionskip}{-0.0cm}   
\setlength{\belowcaptionskip}{-0.0cm} 
\centering
\subfigure[Timer]   
{
  \label{fig:model-timer}%
  \includegraphics[width=0.22\columnwidth]{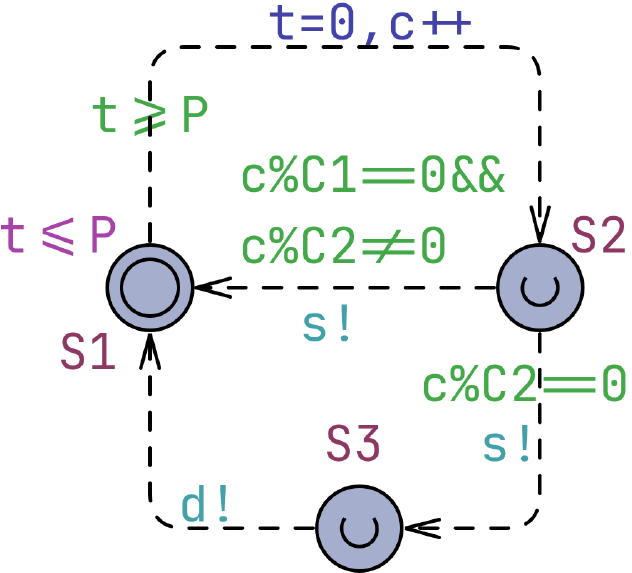}%
}
\subfigure[Controller]   
{
  \label{fig:model-controller}%
  \includegraphics[width=0.27\columnwidth]{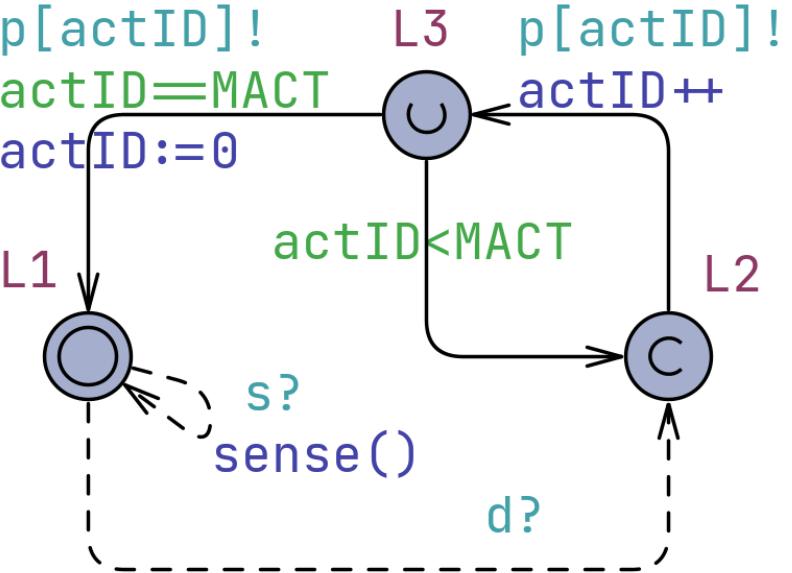}%
}
\subfigure[Action]{
  \label{fig:model-action}%
  \includegraphics[width=0.17\columnwidth]{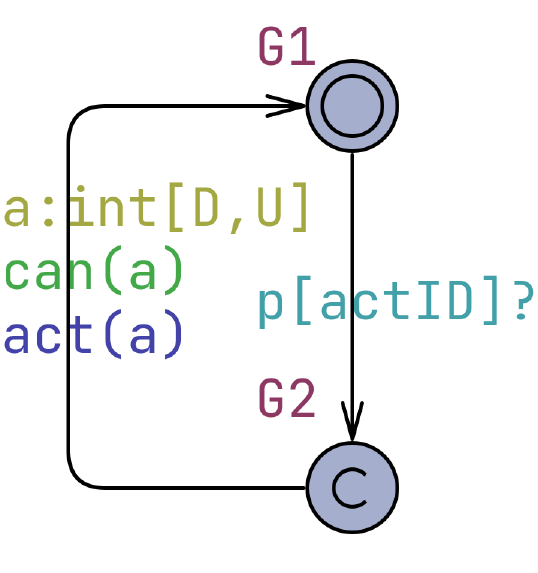}%
}
\subfigure[Dynamics]{
  \label{fig:model-dynamics}%
  \includegraphics[width=0.19\columnwidth]{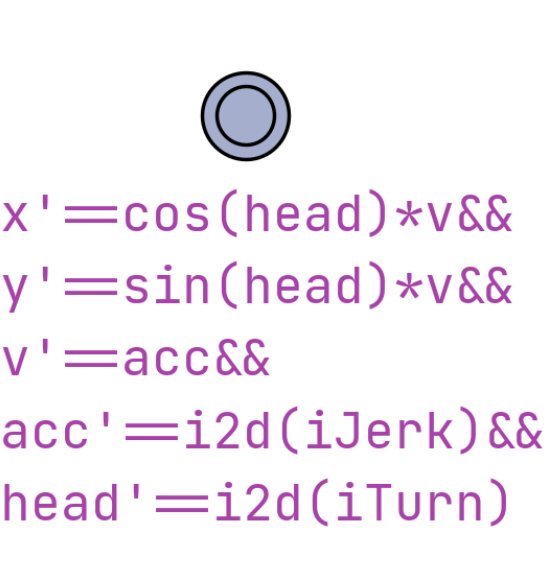}%
}
\caption{UPPAAL model templates of AD vehicles. In (a), (b), and (c), \texttt{s}, \texttt{d}, and \texttt{p} are broadcast synchronisation channels,  \texttt{sense()}, \texttt{can()}, and \texttt{act()} are C-like code functions, \texttt{a}, \texttt{c}, and \texttt{actID} are integers, and \texttt{P}, \texttt{C1}, \texttt{C2}, \texttt{D}, \texttt{U}, and \texttt{MACT} are constant integers, and \texttt{t} is a clock. In (d), variables \texttt{x}, \texttt{y}, \texttt{v}, \texttt{acc}, and \texttt{head} are hybrid clocks, and \texttt{iJerk} and \texttt{iTurn} are integers, equations like \texttt{v'==acc} define the derivatives of the hybrid clocks, and \texttt{i2d} is a C-like code function.}\label{fig:AD-model}
\end{figure}
Fig.\ref{fig:AD-model} depicts three UPPAAL model templates of the AD vehicles, i.e., controller, action, and dynamics.
AD vehicles are cyber-physical systems that consist of digital and physical components.
Fig.\ref{fig:model-timer} - Fig.\ref{fig:model-action} are the digital components describing the control logic.
\textbf{Timer} is a timed automaton calling other models, e.g., \texttt{Controller}, periodically.
This is for modelling CPU periodically calling the processes of the controlling software and reading sensors.
In Fig.\ref{fig:model-timer}, the model leaves location \texttt{S1} every \texttt{P} time unit and comes to an urgent location \texttt{S2} meaning that the next transition starts immediately when reaching this location.
Transitions from \texttt{S2} have two options: i) going back to \texttt{S1} directly, ii) going back to \texttt{S1} via another urgent location \texttt{S3}.
In case i), the transition is synchronised with \texttt{Controller} on channel \texttt{s} meaning that the \texttt{Controller} reads data from sensors by calling function \texttt{sense()}. 
In case ii), the first transition to location \texttt{S3} is the same as case i), which is followed by a transition to location \texttt{S1} synchronised on channel \texttt{d} meaning that the \texttt{Controller} starts to make a decision of actions after reading the sensor data.
Note that locations \texttt{S2} and \texttt{S3} are both urgent, meaning that the operations of sensor-reading, decision-making, and performing actions happen instantaneously at the end of each period. 
Constants \texttt{C1} and \texttt{C2} are for distinguishing the periods of cases i) and ii), which are also used in the model pre-analysis before running RL (Section \ref{sec:preanalysis}).
\textbf{Controller} and \textbf{Action} are also timed automata.
When \texttt{Timer} invokes sensor-reading, \texttt{Controller} transits via a self-loop edge of its initial location \texttt{L1}, meaning that \texttt{Controller} does nothing but reads data from sensors in this period.
When \texttt{Timer} invokes decision-making, \texttt{Controller} transits to a committed location \texttt{L2}, meaning that the next transition must start from this location immediately.
Next, \texttt{Controller} goes to an urgent location \texttt{L3} synchronising on channel \texttt{p[actID]}, in which \texttt{actID} is an integer identifying actions.
In Fig.\ref{fig:model-action}, \texttt{Action} transits to a committed location \texttt{G2} synchronising on channel \texttt{p[actID]} too.
The difference is that now location \texttt{G2} is committed, so transitions from \texttt{L3} in \texttt{Controller} must wait until the ones in \texttt{Action} finish, meaning that actions are atomic and cannot be interrupted. 
From \texttt{G2}, \texttt{Action} goes back to \texttt{G1} via multiple actions although only one edge appears from \texttt{G2} to \texttt{G1}.
The author labels this edge with a \texttt{select} statement, which assigns different integers to an edge.
In this way, an edge can represent multiple transitions at the semantic level.
Here, \texttt{a:int[D,U]} means an integer \texttt{a} from constant \texttt{D} to constant \texttt{U} is selected, where \texttt{a} is a variable associated to this action, and \texttt{D} and \texttt{U} are the lower and upper boundaries of
\texttt{a}, respectively.
This design is for modelling continuous actions.
In the field of control theory, real numbers are often used to present continuous actions and their dynamics.
However, it is impossible to represent real numbers in digital systems precisely, because digital systems use floating-point numbers to represent real numbers.
A truncation is inevitable when the real number is not rational or its digit lengths exceed the limit of the digital system, e.g., 32 bits or 64 bits.
In the model-checking world, floating-point numbers are not welcome as they would overly bloat a model's state space and floating-point computations are unstable due to the \textit{cancellation} effects \cite{ivanvcic2010numerical}.
For example, if a model contains floating-point numbers (aka, type \texttt{double}), UPPAAL does not allow symbolic analysis, e.g., symbolic model checking and ESS.
To avoid this problem, floating-point numbers in the new model templates are either abstracted away from the symbolic state space, e.g., hybrid clocks, or represented as integers.
Specifically, the \textit{base} and \textit{exponent} of floating-point numbers are defined as constant integers, and the \textit{significand} of each floating-point number is an integer, which is used to represent this floating-point number.
For example, when the base is $10$ and the exponent is $-4$, $21200$ represents $2.12$.
\begin{equation}\label{eq:floating-point}
        2.12 = \underbrace{21200}_\textit{significand} \times \overbrace{\underbrace{10^{-4}}_\textit{base$^\text{exponent}$}}^\textit{scale}
\end{equation}
In the model \texttt{Action}, constants \texttt{D} and \texttt{U} are the integer representations of two floating-point numbers, that is, the lower and upper boundaries of continuous variable \texttt{a}.
When the period of decision-making comes, \texttt{Action} needs to choose a value from \texttt{D} to \texttt{U} for variable \texttt{a}, which indicates selecting a continuous action.
One can set the granularity of continuous actions by changing the exponent of floating-point numbers.
Although this representation is an approximation of real numbers, it is how digital systems work and allows symbolic analysis in UPPAAL.
\textbf{Dynamics} is a hybrid automaton (Fig.\ref{fig:model-dynamics}), where variables \texttt{x}, \texttt{y}, \texttt{v}, \texttt{acc}, and \texttt{head} are hybrid clocks, indicating the AD vehicle's X and Y coordinates, velocity, acceleration, and heading, respectively.
Equations, such as \texttt{v'==acc}, define the derivatives of the hybrid clocks.
The rate of acceleration (aka, jerk) and turning are modelled as integers (i.e., \texttt{iJerk} and \texttt{iTurn}) because these two variables are changed by actions, and the integers are the significands of the floating-point numbers associated to continuous actions.
Therefore, before they are used in any computation, such as the equations of derivatives, they must be transformed back to floating-point numbers.
This is done in function \texttt{i2d}, where the significand multiplies the scale and becomes a floating-point number.
Oppositely, another function \texttt{d2i} transforms a floating-point number into an integer by dividing the former by the scale.
\subsection{Numerical Integration}
Since the continuous variables in \texttt{Dynamics} are hybrid clocks, one cannot use them in symbolic analysis in UPPAAL.
However, functions like \texttt{sense()} are supposed to detect the values of these variables.
Thanks to the integer representations of floating-point numbers, the author implements a function for calculating the values of continuous variables by using numerical integration.
Algorithm \ref{algo:numerical-integration} shows the numerical integration in function \texttt{sense()}.
\begin{algorithm}[H]
\small 
\setstretch{0.5}
\caption{sense(): numerical integration}\label{algo:numerical-integration}
\DontPrintSemicolon
\SetKwFunction{FUpdateDiscrete}{sense}
\SetKwProg{Fn}{Function}{}{}
\Fn{\FUpdateDiscrete{}}{
    int steps := 1/G\tcp*{G is the granularity of integration}
    double x := \texttt{i2d}(iX)\tcp*{convert integer iX to double x}\label{code:integerToDouble-start}
    double y, v, acc, head ...\tcp*{convert the rest integers to double}\label{code:integerToDouble-end}
    double unit = C1 $\cdot$ G\tcp*{C1 is defined in Timer, Fig.\ref{fig:model-timer}}
    \While{steps $\neq$ 0}
    {
        acc := acc + iJerk $\cdot$ unit\;\label{code:while-start}
        head := head + iTurn $\cdot$ unit\;
        v := v + acc $\cdot$ unit\;
        x := x + v $\cdot$ \texttt{cos}(head) $\cdot$ unit\; 
        y := y + v $\cdot$ \texttt{sin}(head) $\cdot$ unit\; 
        steps := steps - 1\;\label{code:while-end}
    }
    iX := \texttt{d2i}(x)\tcp*{convert double x to integer iX}\label{code:doubleToInteger-start}
    iY, iV, iAcc, iHead ...\tcp*{convert the rest double to integers}\label{code:doubleToInteger-end}
    \Return\;
}
\end{algorithm}
\setlength{\textfloatsep}{0pt}
Line \ref{code:integerToDouble-start} and line \ref{code:integerToDouble-end} convert variables like \texttt{iX} to floating-point numbers.
From line \ref{code:while-start} to line \ref{code:while-end}, the results of the ODE in the model \texttt{Dynamics} are approximated by the numerical integration.
Specifically, the floating-point numbers are updated step by step with a changing unit $C1 \cdot G$, where $C1$ is the length of the sensing period and $G$ is a predefined sampling granularity of the integration points.
Although the numerical integration only approximates the integral, it reflects what the digital system perceives via sensors, that is, periodically updated discrete variables. 
Between two consecutive sampling periods, the variables' values remain unchanged in the controller.
After the numerical integration, line \ref{code:doubleToInteger-start} and line \ref{code:doubleToInteger-end} convert the floating-point numbers back to their integer representations, which can be used in functions like \texttt{collide()} and \texttt{offroad()}.
\subsection{Model Template of Moving Obstacles}
\begin{wrapfigure}[9]{l}{0.3\columnwidth}
\vspace{-.5cm}
\setlength{\abovecaptionskip}{-0.00cm}   
\setlength{\belowcaptionskip}{-0.00cm} 
\begin{center}
  \includegraphics[width=0.28\columnwidth]{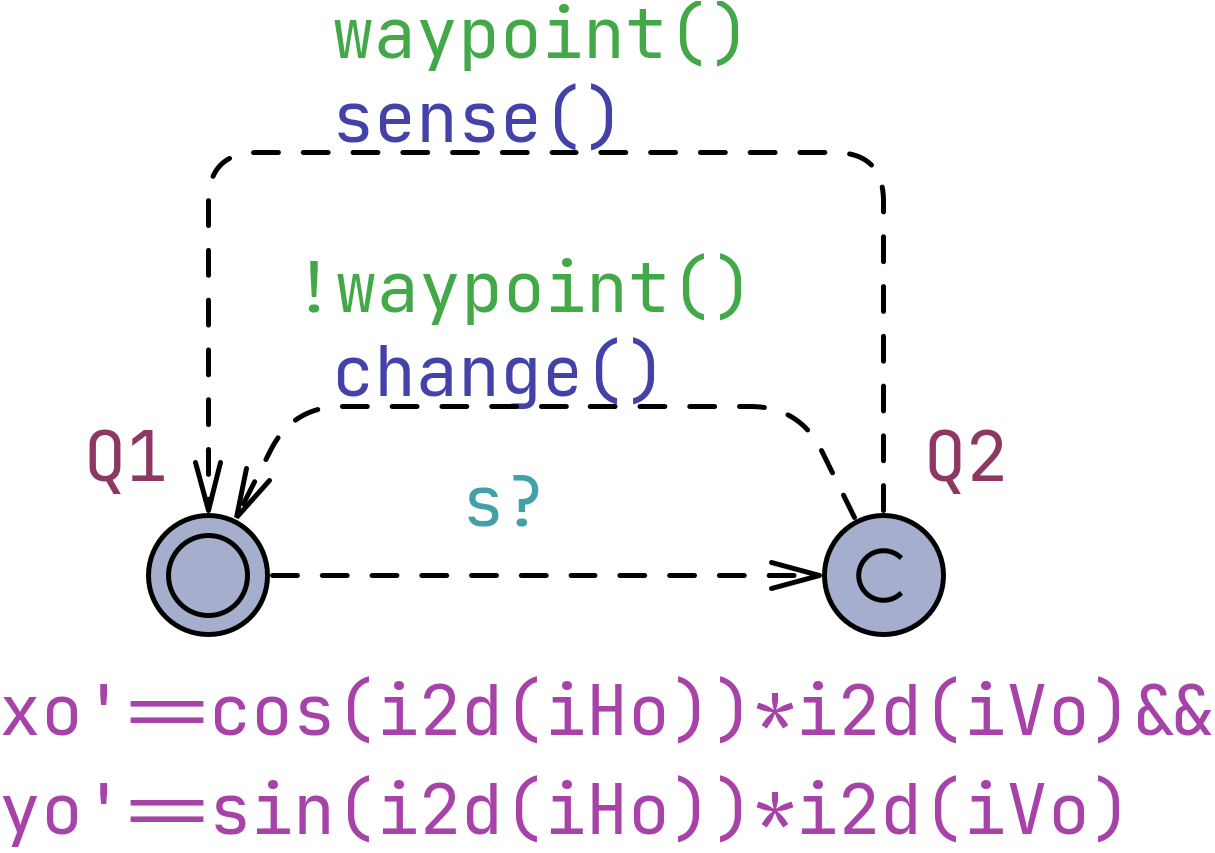}
  \caption{Obstacle}
  \label{fig:model-obstacle}
\end{center}
\end{wrapfigure}
Moving obstacles in CommonRoad usually follow predefined trajectories (Fig.\ref{fig:commonupproad}), therefore the model template of moving obstacles is relatively simple.
Fig.\ref{fig:model-obstacle} is the model template, where \texttt{xo} and \texttt{yo} are the only two hybrid clocks because the velocity (\texttt{iVo)} and heading (\texttt{iHo}) of a moving obstacle immediately changes when it reaches a waypoint on the trajectory.
One may argue that the model template is not faithful to real moving obstacles because the velocity and heading of an object in the real world must be continuous.
However, this model template is based on two conditions: i) the moving obstacle follows a predefined trajectory, and ii) the controller of the AD vehicle is digital.
Condition i) is explained, so the moving obstacle can only change its velocity and head at the waypoints of the predefined trajectory.
Condition ii) is also true because our AD vehicle is a cyber-physical system.
Therefore, the information on moving obstacles is discrete from the AD vehicle's point of view.
Besides, this modelling removes unnecessary details, which simplifies verification and synthesis.
One can easily add probabilistic behaviour to moving obstacles, like the uncertain velocity due to erroneous sensors of AD vehicles.
During learning and statistical model checking, non-deterministic transitions in the model template are interpreted as stochastic transitions with a uniform probability distribution by UPPAAL.
However, this paper does not focus on such behaviours and interested readers are referred to the website of CommonUppRoad\footnote{sites.google.com/view/commonupproad}.
\section{Model Checking for AD Reinforcement Learning}\label{sec:mc4rl}
In this section, the author introduces three aspects that model checking (MC) can do for reinforcement learning (RL) in the context of AD vehicles.
First, MC enables model pre-analysis and quick checking for the existence of an optimal motion plan.
Second, to achieve the best learning performance, MC helps to choose a suitable decision-making period, construct reward automata, and synthesise a safety shield for the learning process.
Last, MC can verify the learning results.
\subsection{Pre-analysis of AD Vehicle and Scenario Models}\label{sec:preanalysis}
\begin{figure}[t]
\vspace{-0.0cm}
\setlength{\abovecaptionskip}{-0.0cm}   
\setlength{\belowcaptionskip}{+0.5cm} 
\centering
\subfigure[Lane changing]   
{
  \label{fig:lanechange-normal}%
  \includegraphics[width=0.2\columnwidth]{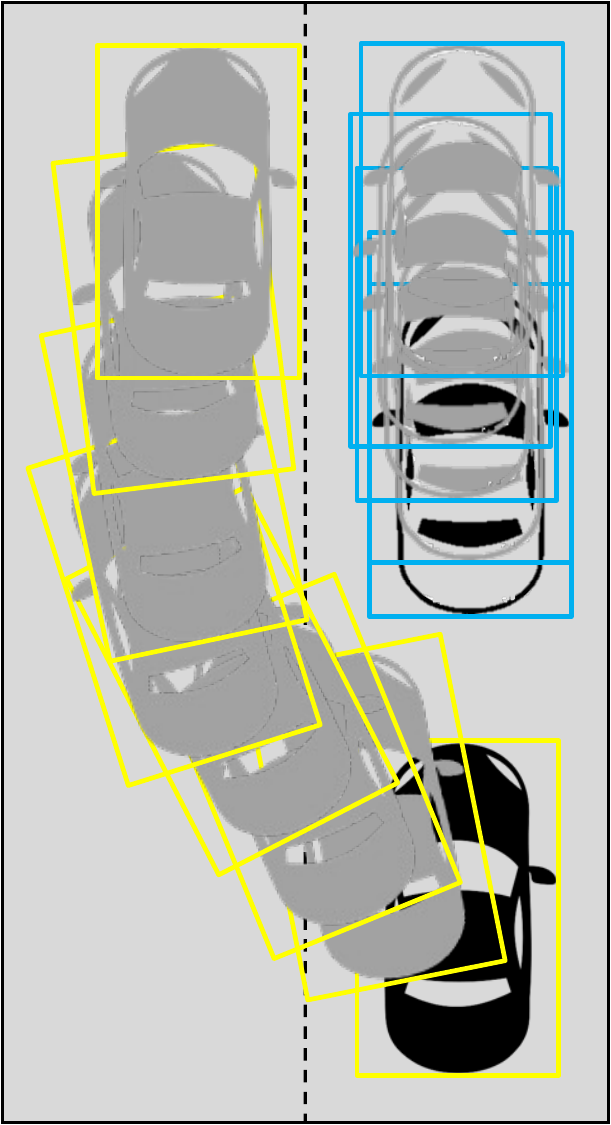}%
}
\subfigure[A missing frame]   
{
  \label{fig:lanechange-collision}%
  \includegraphics[width=0.2\columnwidth]{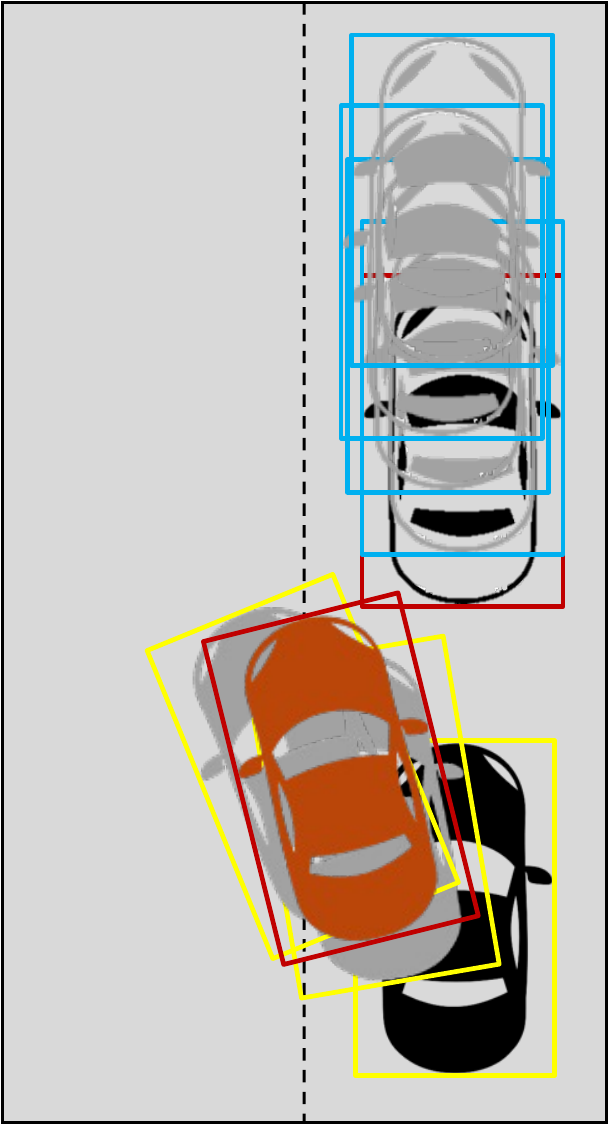}%
}
\subfigure[No valid motion plan]{
  \label{fig:lanechange-obstacle}%
  \includegraphics[width=0.2\columnwidth]{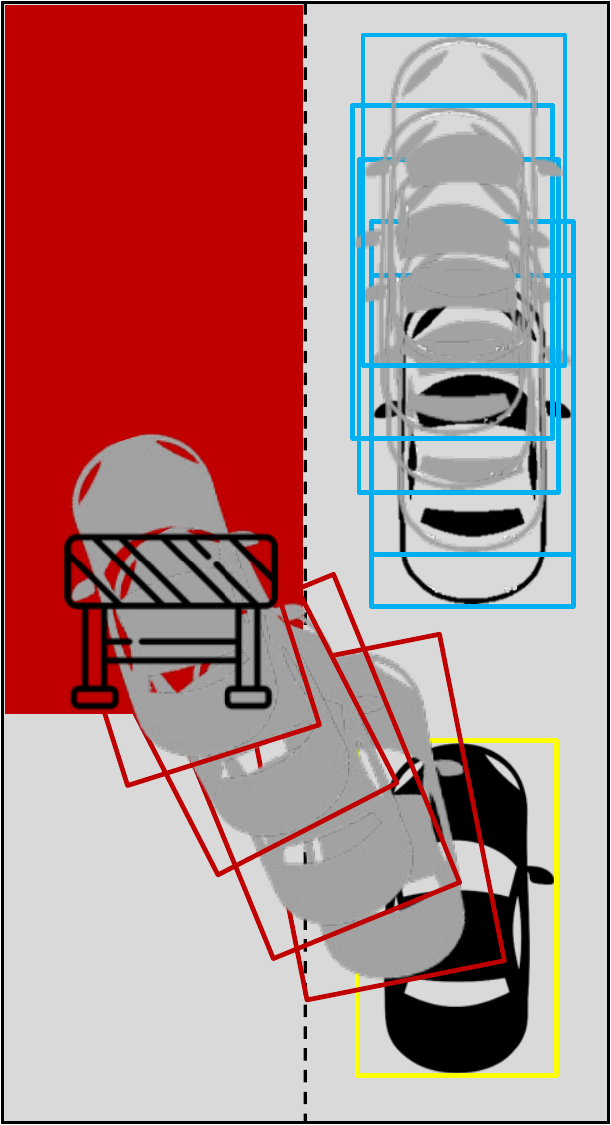}%
}
\caption{An illustration of potential problems when running RL without pre-analysis.}\label{fig:lanechange}
\end{figure}

When facing a motion-planning problem such as the one in Fig.\ref{fig:lanechange}, the first task is probably designing a good searching algorithm to find the optimal state-action pairs in the model state space.
However, such solutions neglect two important aspects that should be considered before the motion planning starts.
First, all AD vehicles are cyber-physical systems, which means the controller is a piece of software that periodically reads sensor data, makes decisions, and sends signals to actuators to control the physical processes of the vehicle.
Between two periods, the vehicle moves according to the latest controlling signals.
For example, Fig.\ref{fig:lanechange-normal} depicts the discrete frames capturing the AD vehicle's trajectory, which is safe because the yellow boxes do not overlap with the blue boxes, i.e., the collision detection ranges of the AD vehicle and moving obstacle, respectively.
However, such a trajectory does not guarantee a safe lane-changing manoeuvre.
If the sensing periods are too long, the AD vehicle may miss critical frames of collision (Fig.\ref{fig:lanechange-collision}).
If the decision-making periods are too short, learning may spend too much time on trivial behaviour.
Additionally, computer systems inevitably lose accuracy when representing real numbers.
The mismatch of data types would result in unexpected behaviour.

Fig.\ref{fig:lanechange-obstacle} shows another scenario where running RL is meaningless because a safe motion plan does not even exist. 
Therefore, one may want to ensure the existence of an optimal motion plan before consuming any resource for learning.
However, checking the existence of an optimal motion plan can be as difficult as finding one, because motion planning usually has multiple objectives, such as safety, progress, comfort, and traffic rules conformance.
However, it is relatively easy to decide the non-existence of valid motion plans because safety is the first and foremost target of motion planning.
If one can quickly conclude the non-existence of safe motion plans, one does not even need to run RL.
Symbolic model checking is a powerful tool for this job.
\begin{table}[htp]
\centering
\caption{UPPAAL queries for model pre-analysis, where \texttt{fabs} returns the absolute value of a floating-point number, \texttt{cv} is a hybrid clock and \texttt{iv} is an integer, \texttt{THD} is the threshold of sensor errors.}
\label{tab:preanalysis}
\resizebox{0.99\textwidth}{!}{%
\begin{tabular}{|l|l|l|}
\hline
 & \multicolumn{1}{c|}{Query} & \multicolumn{1}{c|}{Explanation} \\ \hline
$Q_{a}$ & \texttt{Pr[<=MAXT]\{fabs(cv-i2d(iv))$\geq$THD\}} & Probability of sensor errors \\ \hline
$Q_{b}$ & \texttt{E[]!collide()$\&\&$!offroad()} & Existence of a safe path \\ \hline
$Q_{c}$ & \texttt{E<>!collide()$\&\&$!offroad()$\&\&$goal()} & Existence of a safe and reachable path \\ \hline
$Q_{d}$ & \texttt{strategy safe=control:A[] !collide()$\&\&$!offroad()} & Safety shield for RL \\ \hline
\end{tabular}%
}
\end{table}
As aforementioned in Section \ref{sec:model-templates}, the controller model has two periods, one for reading data from sensors and one for decision-making.
One can define the lengths of those two periods and execute the UPPAAL queries in Table \ref{tab:preanalysis} for model pre-analysis.
In $Q_{a}$, \texttt{cv} is a hybrid clock representing a continuous variable in the physical component of the AD vehicle, i.e, model \texttt{Dynamics} in Fig.\ref{fig:model-dynamics}, and \texttt{iv} is an integer used in the controlling software, i.e., model \texttt{Controller} in Fig.\ref{fig:model-controller}.
When the difference between these two variables exceeds a threshold, it means the information perceived by the controlling software is too far away from the ground truth, and thus a sensor error is discovered.
$Q_{a}$ uses statistical model checking and returns the probability of sensor errors occurring.
For example, when \texttt{cv} represents the physical distance between the AD vehicle and the front car, \texttt{iv} represents the same distance but calculated by using discrete frames, and the difference between \texttt{cv} and \texttt{iv} is too large, it indicates the sensing periods are too coarse and a colliding scenario similar to Fig.\ref{fig:lanechange-collision} may exist.
$Q_{b}$ and $Q_{c}$ use symbolic model checking, which exhaustively explores the entire symbolic state space of the model.
If $Q_{b}$ is satisfied, an absolutely safe path exists in the state space where no collision or offroad event happens, otherwise, no safe path exists and thus RL is not needed any more.
However, the path found by $Q_{b}$ does not necessarily reach the goal.
$Q_{c}$ adds a condition \texttt{goal()}, which returns \textit{true} when the AD vehicle reaches the destination.
If $Q_{c}$ is satisfied, a path reaching the destination safely is found, otherwise, a safe and reachable path does not exist and thus RL is not needed either (Fig.\ref{fig:lanechange-obstacle}).
$Q_{d}$ is for synthesising a safe strategy that guarantees the AD vehicle is safe regardless of how other vehicles move in the environment.
This is the same query for synthesising safety shields for RL, i.e., Query (\ref{query:safe}).
$Q_{b}$ - $Q_{d}$ are symbolic analysis, which excludes hybrid clocks. Hence, if a sensor error is indicated by $Q_{a}$, $Q_{b}$ - $Q_{d}$ are not needed because even if a symbolic safe path and a shield are found, they are not necessarily safe in real scenarios.
If $Q_{a}$ returns a value lower than the tolerant level of sensor errors, but $Q_{b}$ or $Q_{c}$ is not satisfied, one may want to shorten the decision-making periods, change or add the AD actions, or extend the time limit.
All of these changes can be easily configured in the Python code of CommonUppRoad.
One does not need to know the formal model templates (Fig.\ref{fig:AD-model}) behind the Python code, which makes CommonUppRoad user-friendly to researchers outside the FMs community.
Now if $Q_{a}$ - $Q_{c}$ all show positive results but $Q_{d}$ fails, it means the moving obstacles, like the front car in Fig.\ref{fig:lanechange}, can make the AD vehicle deviate from the path of $Q_{b}$ or $Q_{c}$.
One can modify the behaviour of the AD vehicle or shorten the decision-making periods so that the AD vehicle has more chances to avoid moving obstacles.
In some cases, however, the AD vehicle only needs to make decisions at a few critical time steps, but RL with a fixed learning step size makes it unnecessarily complex. 
Having different lengths of sensing and decision-making periods makes RL in CommonUppRoad adaptable to simple and complex scenarios.
One can start with a large decision-making period but a standard sensing period, and run $Q_{d}$ to see if a safety shield exists.
If it does exist, one can decompose the motion-planning problem into sub-problems by dividing it at the time points of decision-making.
%
%
%
Next, one can fine-tune the motion plan for each of the sub-problems by using RL.
We refer interested readers to our previous work about this concatenated motion planning and verification \cite{naeem2024energy}.
\subsection{Model Checking Facilities Multi-Objective Learning}
Autonomous driving usually involves multiple objectives. Naturally, safety and reachability are two major objectives, which require the AD vehicle to reach the destination without collision or going off the road.
Additionally, comfort and traffic rule conformance are also important objectives of AD.
In some applications, timing and efficiency are also non-negligible requirements.
Although the objectives of AD are multiple, the goal of RL is simple, i.e., maximising the cumulative reward.
Therefore, the design of AD reward functions must take into account all the objectives, which is not trivial because those objectives can be contradictory.
For example, RL rewards AD vehicles for progressing towards the goal and punishes them for collisions.
When a permanent obstacle blocks the only path to the goal, RL may eventually motivate the AD vehicle to crash into the obstacle if the cumulative waiting penalty exceeds the collision penalty.
This erroneous behaviour stems from AD engineers' insufficient understanding of reward functions in the AD context.
Formal models, however, provide rigorous semantics of the AD vehicles' behaviour as well as reward functions, which would greatly help the AD engineers design multi-objective reward functions and even verify them before running RL.
In this paper, the author selects three AD objectives in the literature \cite{abouelazm2024review}, that is, safety, progress, and comfort, and proposes different methods to cope with them.
Additionally, the reward functions of all objectives are integrated into one automaton, which is similar to the concept of \textit{reward machines} in the literature \cite{icarte2022reward}.
This design has many benefits. First, it enables the users of CommonUppRoad to consider the various driving contexts in one model. Under different contexts, e.g., different weather conditions, the reward functions can be different.
Second, a reward automaton allows users to verify the design of reward functions before RL.
One can observe the change of rewards by exploring the model's symbolic state space step by step, or model check properties, such as the penalty for unsafe behaviour is always larger than the summation of rewards of other objectives.
\subsubsection{Reward Automata}
\begin{wrapfigure}[20]{l}{0.5\columnwidth}
\vspace{-.7cm}
\setlength{\abovecaptionskip}{-0.00cm}   
\setlength{\belowcaptionskip}{-0.00cm} 
\begin{center}
  \includegraphics[width=0.47\columnwidth]{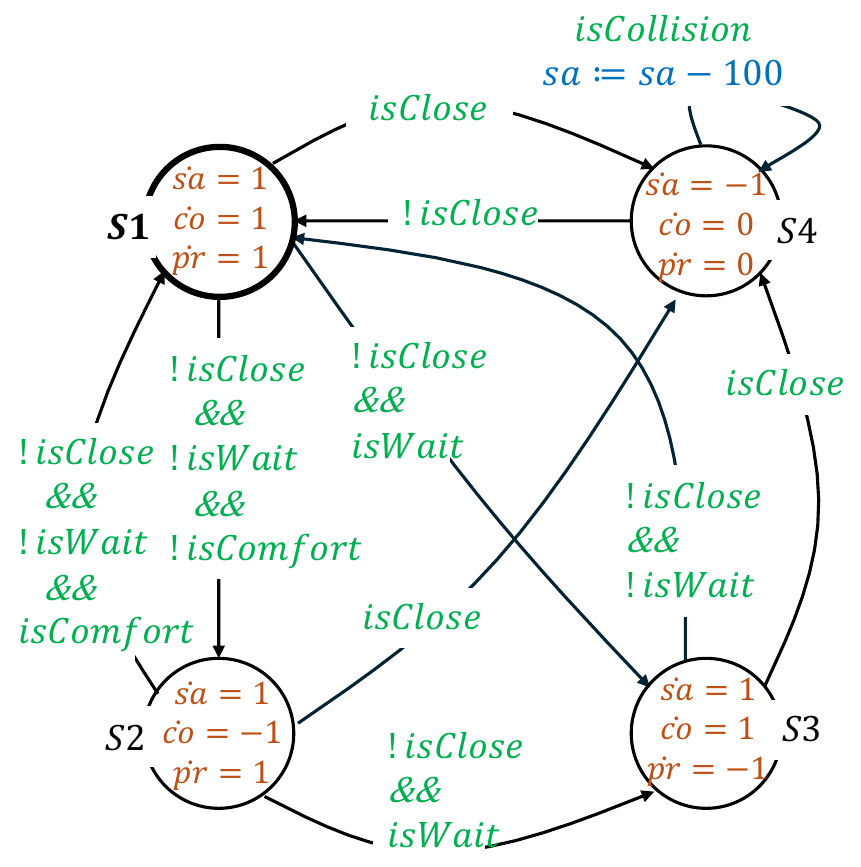}
  \caption{An example of reward automata. Hybrid clocks \textit{sa}, \textit{co}, and \texttt{pr} represent the rewards for safety, comfort, and progress, respectively.}  \label{fig:reward-automaton}
\end{center}
\end{wrapfigure}
Fig.\ref{fig:reward-automaton} shows an example of reward automata, which considers the aforementioned three objectives in the following order of priority: \textit{safety}$>$\textit{progress}$>$\textit{comfort}.
State \texttt{S1} is the initial state, where the AD vehicle moves normally and the three rewards increase steadily at the rate of one.
Once the distance between the AD vehicle and an obstacle is lower than a threshold, i.e., \texttt{isClose} is \textit{true}, the reward automaton transits to state \texttt{S4}, where \texttt{sa} decreases and the other two rewards stay the same.
This indicates that once an unsafe situation occurs, no reward should increase.
This prohibits the cumulative rewards from becoming larger than the penalty for unsafe behaviour.
Similarly, the reward automaton can transit to states \texttt{S2} and \texttt{S3}, where the rewards for comfort and progress decrease, respectively.
State \texttt{S4} has a self-loop, where a collision happens and \texttt{sa} is divided by a large value, e.g., 100.
This design follows the recommendation from the literature \cite{abouelazm2024review}, i.e., a safety reward should include a sparse penalty for collisions and a continuous dense term that penalises dangerous behaviour.
However, if one synthesises a safety shield for RL first ($Q_{d}$ in Table \ref{tab:preanalysis}), this self-loop is not necessary because the safety shield eliminates all collisions and offroad events.
One can verify reward automata together with the AD vehicle model, which helps in understanding and improving reward functions.
Table \ref{tab:reward-automata} lists some exemplary UPPAAL queries for reward automata verification.
$Q_{e}$ - $Q_{g}$ verify if the reward automaton transits to the right state when the AD vehicle's behaviour presents the corresponding features.
For example, $Q_{g}$ verifies whether the reward automaton goes to the right state to decrease the comfort reward when the acceleration or the angular speed becomes larger than $80\%$ of its maximum value, meaning the AD vehicle is accelerating or turning too much and makes passengers feel uncomfortable.
However, the reward automaton only does this transition when the Boolean variable \texttt{comf}, i.e., comfort, is false, and other Boolean variables \texttt{prog} and \texttt{safe} are true, because the objective \textit{comfort} has the lowest priority among all objectives.
$Q_{f}$ is designed similarly.
%
%
Counterexamples returned from these queries greatly help the AD vehicle designers to see the problems of their reward functions and improve them before RL starts.
\begin{table}[htp]
\centering
\caption{Examplary UPPAAL queries for the verification of reward automata \texttt{RA}, where \texttt{safe}, \texttt{prog}, and \texttt{comf} are Boolean variables, \texttt{iX} is the integer representation of a hybrid clock \texttt{x}, \texttt{TD} is the distance threshold of collision, \texttt{S} is a non-negtive integer, \texttt{MA} and \texttt{MT} are maximum acceleration and turning speed, respectively, $p_1$ and $p_2$ are real numbers between 0 and 1, \texttt{AD$_N$} and \texttt{Obs$_N$} are the positions of the AD vehicle and an obstacle at the N$_{th}$ period, respectively, \texttt{GO} is the goal area, \texttt{Dis($\Box_1$,$\Box_2$)} computes the distance between two rectangles, which represent objects in the environment, and \texttt{w1} - \texttt{w3} are weights.}
\label{tab:reward-automata}
\resizebox{1\textwidth}{!}{%
\begin{tabularx}{\textwidth}{|l|X|X|}
\hline
 & \multicolumn{1}{c|}{Query} & \multicolumn{1}{c|}{Explanation} \\ \hline
 $Q_{e}$ & \texttt{A[] !safe imply RA.S4, safe=Dis(AD$_N$,Obs$_N$)$> S\cdot TD$} & If AD vehicle and an obstacle gets too close, the safety reward is punished. \\ \hline
$Q_{f}$ & \texttt{A[] !prog\&safe imply RA.S3, prog=Dis(AD$_N$,GO)$>$Dis(AD$_{N+1}$,GO)} & If the distance from AD to the goal increases, the progress reward is punished. \\ \hline
$Q_{g}$ & \texttt{A[] !comf\&prog\&safe imply RA.S2, comf=iAcc$<p_1\cdot$MA\&iTurn$<p_2\cdot$MT} & If acceleration or angular velocity is too large, the comfort reward is punished. \\ \hline
$Q_{h}$ & \texttt{A[] RA.S4 imply w1$\cdot$iSa$+$w2$\cdot$iCo+w3$\cdot$iPr$\leq$0} & The safety reward cannot be compensated by other rewards when unsafe behaviour occurs. \\ \hline
\end{tabularx}%
}
\end{table}
\section{Experiments}\label{sec:experiment}
\begin{wrapfigure}[12]{l}{0.35\columnwidth}
\vspace{-.5cm}
\setlength{\abovecaptionskip}{-0.00cm}   
\setlength{\belowcaptionskip}{-0.00cm} 
\begin{center}
  \includegraphics[width=0.35\columnwidth]{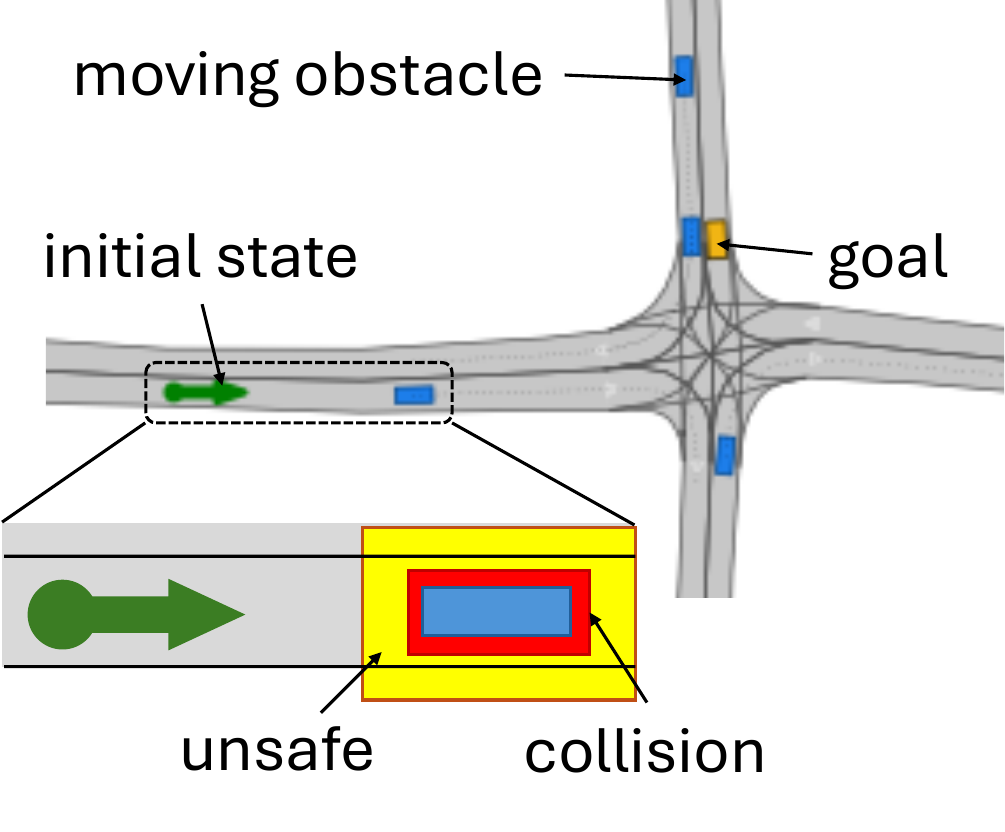}
  \caption{Experimental AD scenario}  \label{fig:experiment-scenario}
\end{center}
\end{wrapfigure}
The experiments in this section aim to demonstrate the strengths of MC-enhanced RL in two aspects: i) the necessity of model pre-analysis, and ii) the benefits of MC in designing reward automata\footnote{The model for experiments: sites.google.com/view/commonupproad/experiment. Run it in UPPAAL 5.1.0-beta5.}.
The selected AD scenario is depicted in Fig.\ref{fig:experiment-scenario}, where the AD vehicle needs to enter the intersection, turn left, and drive to the goal area safely.
Surrounding obstacles, there are two critical ranges: \textit{unsafe} and \textit{collision}.
Once the distance between the AD vehicle and an obstacle is less than \texttt{TD} (resp. 3$\cdot$\texttt{TD}), a collision (resp. unsafe) situation happens.
To show the necessity of model pre-analysis, the author intentionally decreases the sensor accuracy and extends the sensing periods.
Specifically, the exponent (see Equation (\ref{eq:floating-point})) is decreased from four to one such that the integer representations of floating-point numbers preserve only one digit after the decimal.
The sensing periods are set to be two time units such that sensors may miss critical frames.
First, the author executes Query (\ref{query:safe}) to synthesise a safety shield, namely \texttt{safe}, and verify the model by the following queries.
\begin{equation}\label{query:unsafe-veri-pr}
\texttt{Pr[$\leq$MAXT](<> Dis(AD,Obs)$\geq 3\cdot$TD) under safe}
\end{equation}
\begin{equation}\label{query:unsafe-veri-si}
\texttt{simulate[$\leq$MAXT;100]($3\cdot$TD,Dis(AD,Obs)) under safe}
\end{equation}

\begin{figure}[t]
\setlength{\belowcaptionskip}{+0.5cm} 
\centering
\subfigure[Exponent: 1, sensing period: 1]   
{
  \label{fig:exp-1-N2-1}%
  \includegraphics[width=0.32\columnwidth]{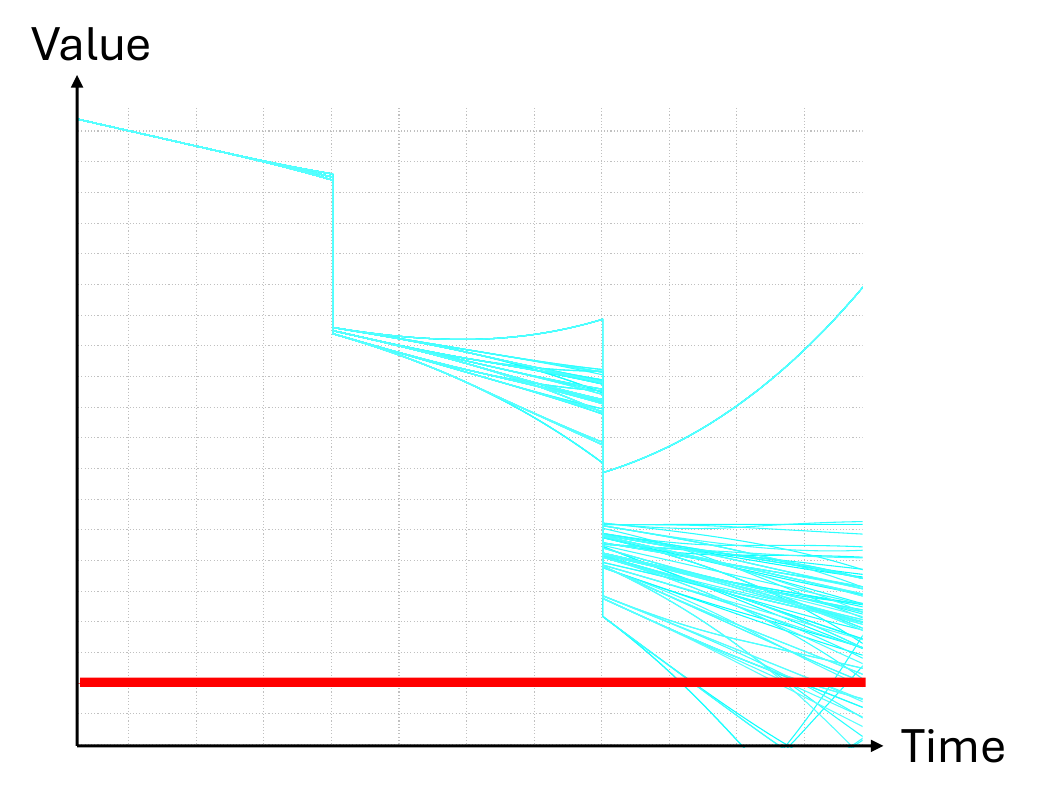}%
}
\subfigure[Exponent: 4, sensing period: 2]   
{
  \label{fig:exp-1-N2-2}%
  \includegraphics[width=0.32\columnwidth]{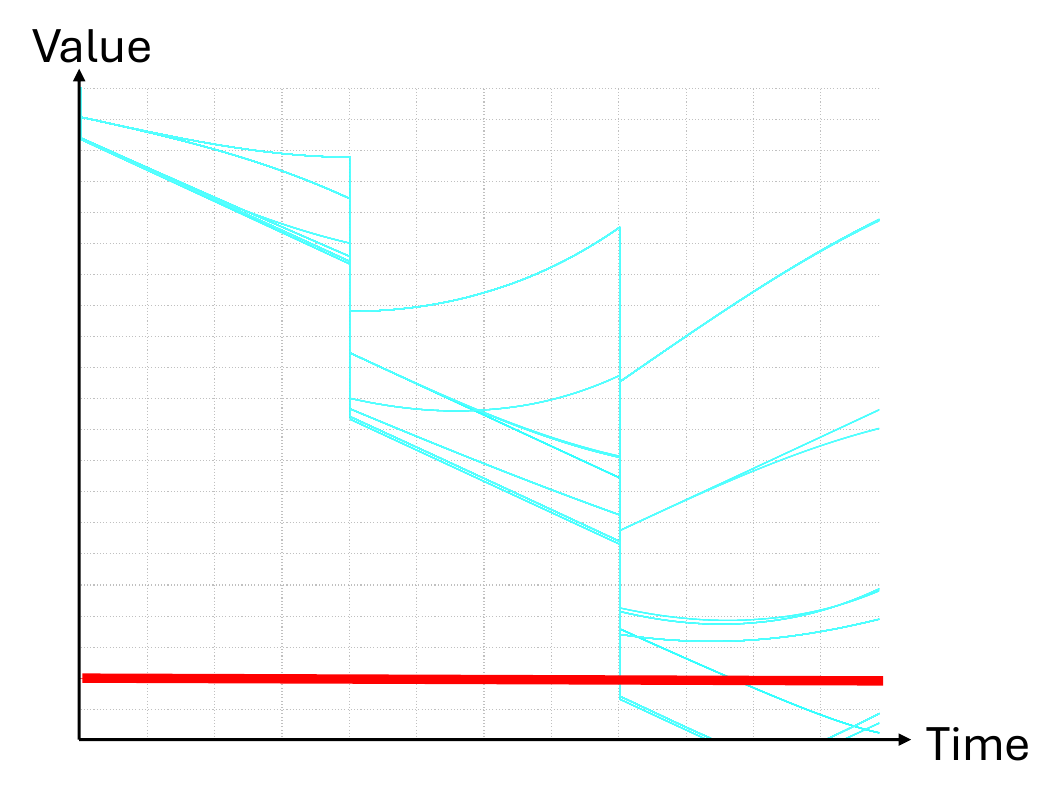}%
}
\subfigure[Exponent: 4, sensing period: 1]{
  \label{fig:exp-4-N2-1}%
  \includegraphics[width=0.32\columnwidth]{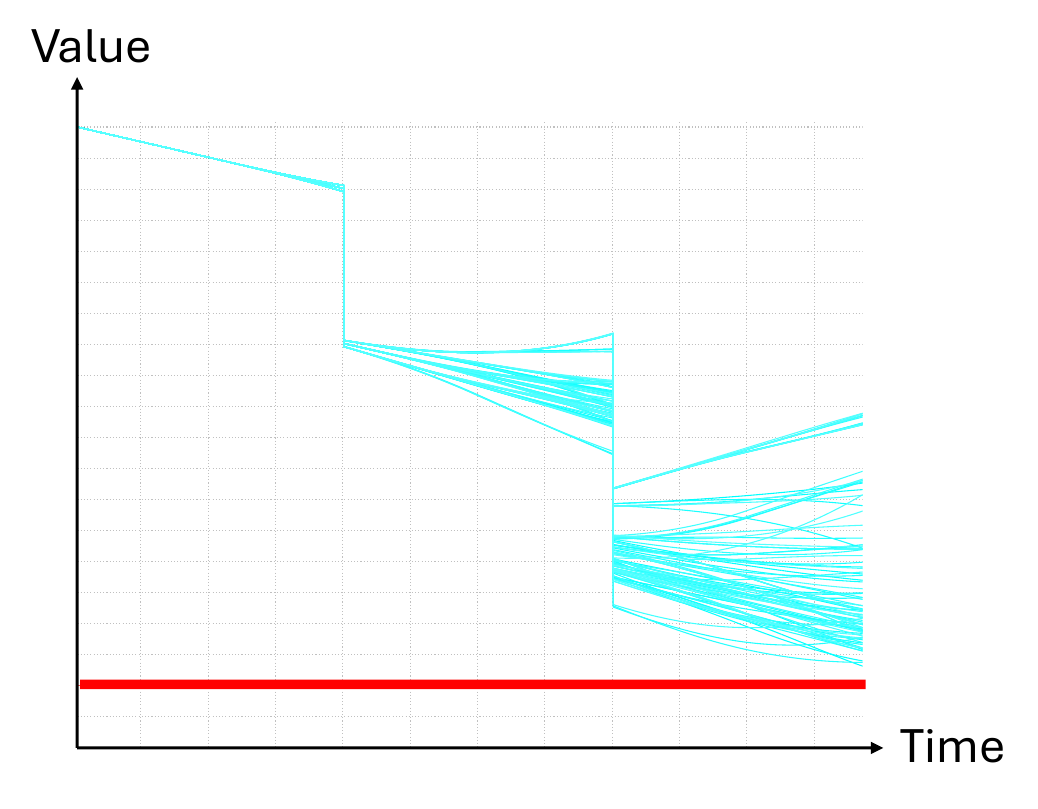}%
}
\caption{Model pre-analysis result (Query (\ref{query:unsafe-veri-si})). The blue lines are the distances between the AD vehicle and an obstacle in 100 simulations. The red line is the threshold of unsafe distance, i.e., 3$\cdot$\texttt{TD}.}\label{fig:experiment-preanalysis}
\end{figure}

Fig.\ref{fig:experiment-preanalysis} shows the results of Query (\ref{query:unsafe-veri-si}) in different settings and 100 simulations.
In the first two settings, the AD vehicle's distance to an obstacle can be less than the threshold even though the model is under the control of the safety shield.
This shows that the exponent and sensing period are inadequate. 
Fig.\ref{fig:exp-4-N2-1} has no cases exceeding the threshold, and thus the author chooses this setting in the following experiments.
There are several RL algorithms implemented in UPPAAL, such as Q learning. One can also call their own RL algorithm in UPPAAL from an external library \cite{gu2022correctness}.
The next experiment uses two synthesis methods: Q learning with/without a safety shield, using a reward automaton or a reward function.
First, the author executes Query (\ref{query:reach}), in which \texttt{dv1} and \texttt{cv1}, etc., are state variables used in RL, and \texttt{rf} can be a reward function (Equation (\ref{eq:reward-function})) or a summation of hybrid clocks (Equation (\ref{eq:reward-automata})), where \texttt{safe}, \texttt{prog}, and \texttt{comf} are Boolean variables (Table \ref{tab:reward-automata}), and \texttt{sa}, \texttt{pr}, and \texttt{co} are hybrid clocks (Fig.\ref{fig:reward-automaton}).
%
%
As strategy \texttt{reach} may not be safe or reach the goal, the author verifies it against Queries (\ref{query:safe-veri-reach}) and (\ref{query:reach-veri-reach}).
\begin{equation}\label{query:reach}
\texttt{strategy reach = maxE(rf)[<=MAXT]\{dv1,...\}->\{cv1,...\}: <> time>=MAXT}
\end{equation}
\begin{equation}\label{eq:reward-function}
\texttt{rf}=10\cdot\texttt{safe}+(5+100\cdot goal())\cdot\texttt{prog}+\texttt{comf}
\end{equation}
\begin{equation}\label{eq:reward-automata}
\texttt{rf}=\texttt{sa}+\texttt{pr}+\texttt{co}
\end{equation}
\begin{equation}\label{query:safe-veri-reach}
\texttt{A[] !collide() \&\& !offroad() under reach}
\end{equation}
\begin{equation}\label{query:reach-veri-reach}
\texttt{A<> goal() under reach}
\end{equation}

In comparison, the author also synthesises a safety shield (Query (\ref{query:safe})) before RL.
%
%
%
The synthesis query now is Query (\ref{query:reachSafe}) such that the learning process and result are guaranteed to be safe.
The computation time is in Table \ref{tab:experiment-reward}, where the column \textit{learning episodes} refers to the number of simulation rounds that RL used in the experiment.
The experiment is conducted on a Macbook Pro with an Apple M2 Pro chip and 16 GB memory. The OS is Sonoma 14.6.1 and UPPAAL's version is 5.1.0-beta5.
Safety-shield synthesis costs around 30 seconds, which is much longer than what the pre-analysis takes (i.e., $Q_{b}$ and $Q_{c}$), that is, within 3 seconds together.
In the cases where safety-shield synthesis takes too long, the pre-analysis would be even more beneficial.
%
%
The computation time of RL with a reward function and without a safety shield (aka, RF) costs the longest time because it does not have a safety shield to restrict the model space and the reward function does not guide the state-space exploration as efficiently as the reward automaton.
The author would like to discuss more about the benefits of using reward automata (RA) and reward functions (RF).
Table \ref{tab:experiment-reward} indicates that the computation time of RL with a safety shield costs a similar time when using RA or RF.
However, the design of RF benefited heavily from RA in the experiment.
For example, the author chose the weights in Equation (\ref{eq:reward-function}) by observing the behaviour of the reward automaton and verifying queries in Table \ref{tab:reward-automata}.
Those queries helped the author understand the mistakes of weights because the priority order of the objectives was broken, e.g., the accumulated rewards of progressing exceeded the punishment of unsafe situations.
In other words, without reward automata, the design of RF would take much longer time in the trial-and-error process.
However, the computation times of queries in Table \ref{tab:reward-automata} were less than 20 seconds in the experiment.
\begin{table}[htp]
\centering
\caption{RL with/without safety shields (SS), using reward automata (RA) or reward functions (RF)}
\label{tab:experiment-reward}
\resizebox{0.65\textwidth}{!}{%
\begin{tabular}{|c|c|c|c|}
\hline
Combination & Queries & Computation time & Learning episodes \\ \hline
\multirow{2}{*}{SS\&RA} & Query (\ref{query:safe}) & 34.8 s & / \\ \cline{2-4} 
 & Query (\ref{query:reachSafe}) & 15.1 s & 20 \\ \hline
\multirow{2}{*}{SS\&RF} & Query (\ref{query:safe}) & 34.8 s & / \\ \cline{2-4} 
 & Query (\ref{query:reachSafe}) & 16.0 s & 20 \\ \hline
RA & Query (\ref{query:reach}) & 15.1 s & 20 \\ \hline
RF & Query (\ref{query:reach}) & \textbf{310.8 s} & \textbf{500} \\ \hline
\end{tabular}%
}
\end{table}
\section{Related Work}\label{sec:related}
In the formal methods (FMs) community, verification of reinforcement learning in autonomous driving (AD) has drawn wide interest.
Khaitan et al.~\cite{khaitan2022state} propose a curriculum learning approach for training a deep reinforcement learning agent. The performance of the curriculum is tested on the task of traversing unsignalised intersections with the CommonRoad framework.
Naumann et al.~\cite{naumann2019safe} propose a motion planning method through probabilistic analysis under occlusions and limited sensor range, and use a real-world scenario with actual existing occlusions in CommonRoad for the validation.
Liu et al.~\cite{liu2023specification} address specification-compliant motion planning for AD vehicles based on set-based reachability analysis with automata-based model checking. The effectiveness of the methods is demonstrated with scenarios from the CommonRoad benchmark suite.
In comparison, this study emphasises FMs' strong support for RL. The MC-based model pre-analysis and reward automata as well as the corresponding verification are the first attempts.
The support of continuous states and actions, and symbolic and statistical model checking in one model are also profoundly beneficial for deepening users' understanding of the target system.
Another usability barrier of FMs in AD systems is the steep learning curve of the mathematics-based techniques.
Researchers have developed tools to overcome this barrier, such as Kronos \cite{bozga1998kronos}, LTSim \cite{blom2010ltsmin}, and SpaceEx \cite{frehse2011spaceex}.
Bersani \textit{et al.} \cite{bersani2020pursue} present PuRSUE (Planner for RobotS in Uncontrollable Environments), which supports users to configure their robotic applications and automatically generate their controllers by using UPPAAL.
Gu et al. \cite{gu2022malta} develop a tool called \textit{MALTA}, which uses UPPAAL as a backend mission planner for AD vehicles under complex road conditions.
These tools mainly suffer from a common problem: state-space explosion.
Alur et al. \cite{alur2018compositional} propose a compositional method for synthesizing reactive controllers satisfying Linear Temporal Logic specifications for multi-agent systems.
Muhammad et al. \cite{muhammad2024energy} also use the concept of compositional planning for synthesising energy-optimal motion plans.  
The new model templates allow different periods for decision-making and sensing, which greatly eases the computational effort for synthesising safety shields and learning.
The experiment results show that the new method performs better than the first version of CommonUppRoad \cite{gu2024commonupproad}, which demonstrates the improvement of the new model templates.
\section{Conclusion}\label{sec:conclusion}
This paper demonstrates how model checking (MC) can strengthen reinforcement learning (RL) in the domain of autonomous driving (AD).
This study is built upon CommonUppRoad, a platform combining CommonRoad and UPPAAL. The author proposes new model templates for AD vehicles, driving scenarios, and reward automata.
The new model templates contain discrete and continuous components and support symbolic and statistical model checking.
Thanks to the new features, the new version of CommonUppRoad proposes model pre-analysis prior to RL and reward automata for designing reward functions.
The Model pre-analysis can help RL designers find bugs in sensor accuracy and determine period lengths, which are imperative for RL.
Reward automata are greatly beneficial for multi-objective RL.
The experiments demonstrate the necessity of model pre-analysis and the profoundly improved performance of RL with safety shields and reward automata.
One of the future works is investigating the possibility of breaking soft rules to achieve important objectives. How reward automata can help in this setting is unknown.
Another direction is scenario generation and critical scenario identification. A combination of MC and RL would be greatly helpful in discovering collision scenarios in a huge database of scenarios.
\section*{Acknowledgments}
The author acknowledges the support of the Swedish Knowledge Foundation via the project SATISFIES - Holistic Synthesis and Verification for Safe and Secure Autonomous Vehicles, grant nr. 20230047.
\bibliographystyle{eptcs}
%

%
\end{document}